\documentclass{opt2022} 

\title[Quantization-Based Optimization: Alternative Stochastic Approximation of Global Optimization]{Quantization-Based Optimization: Alternative Stochastic Approximation of Global Optimization}

\optauthor{%
\Name{Jinwuk Seok} \Email{jnwseok@etri.re.kr}\\
\Name{Chang Sik Cho} \Email{cscho@etri.re.kr}\\
\addr Future Computing Research Division, Artificial Intelligence Research Lab. ETRI, Rep. Korea}

\newtheorem{assumption}{Assumption}
\newtheorem{theorem-opt}{Theorem}
\newtheorem{lemma-opt}[theorem-opt]{Lemma}
\usepackage{pgf}
\SetKwInput{KwInput}{Input}                
\SetKwInput{KwOutput}{Output} 

\usepackage{verbatim}
\usepackage{booktabs}
\usepackage{listings}
\usepackage{multicols}
\begin{document}

\maketitle

\begin{abstract}

In this study, we propose a global optimization algorithm based on quantizing the energy level of an objective function in an NP-hard problem.
According to the white noise hypothesis for a quantization error with a dense and uniform distribution, we can regard the quantization error as i.i.d. white noise. 
From stochastic analysis, the proposed algorithm converges weakly only under conditions satisfying Lipschitz continuity, instead of local convergence properties such as the Hessian constraint of the objective function. 
This shows that the proposed algorithm ensures global optimization by Laplace's condition. 
Numerical experiments show that the proposed algorithm outperforms conventional learning methods in solving NP-hard optimization problems such as the traveling salesman problem.

\end{abstract}

\section{Introduction}
\label{Introduction}

Finding the global optimum in a non-deterministic polynomial-time hardness problem(NP-hard problem) such as the traveling salesman problem (TSP) has been a critical research theme (\cite{Metropolis:1953, Pincus-1970, Holland_1975, Benioff_1980, Nicolis_1977, Feynman_1982, Khachaturyan_1979, Cern_1985}).
Since \citet{SA_83_01} presented the simulated annealing (SA) in the 1980s, researchers have developed and applied various heuristic algorithms to combinatorial optimization problems, including NP-hard problems (\cite{Hamacher_1957, Gidas-1985, Goldberg_1989, Kennedy_1995, Gambardella_1996, Kadowaki_1998, Stochastic_Ta_1999, Hamacher_2006, Santoro_2006, Zhou_2013, Nielsen_2011, Ashley_2016, google-2019, Guilmeau_2021}).
Despite regarding such heuristic algorithms as alternatives to a stochastic optimization technique, the fundamental dynamics of some of these algorithms remain unclear(\cite{Jiang_2007, Clerc_2011, Bonyadi_2017, Ajagekar_2019}).
Those unclear dynamics result in problems such as the selection of suitable hyperparameters for high optimization performance\cite{Bettina_2015}.

In contrast to conventional natural phenomenon-based optimization algorithms, we propose a quantization-based optimization algorithm with a monotonically increasing quantization resolution in this study.
As shown in \cite{Weiss_1979, Boutalis_1989, Ljung_1985}, the main research topic for quantization has been minimizing the effect of the quantization error in signal processing, and this approach is the same for artificial intelligence and machine learning, as described in \cite{Seide_2014, Christopher_2015, Sangil_2019}.
However, if the distribution of the quantization error is sufficiently dense and uniform, we can let the quantization error be white noise, as presented in \citet{Gray:2006}'s paper. 
In addition, \cite{Jimnez_2007} proved that it is i.i.d white noise if components of a dense quantization error vector are asymptotically pairwise independent and distributed uniformly. 
This property is known as the quantization error's white noise hypothesis(WNH).
Accordingly, we can constitute a proper stochastic global optimization algorithm based on the quantization error.
We provide stochastic analysis to prove the proposed algorithm's weak convergence for global optimization based on Laplace's theorem presented by \cite{Lundy_1986, Geman-1986, Hwang-1987, Locatelli_1996, Seok-2022}, and we apply the proposed algorithm to the TSP to validate the algorithm'. 
In particular, in the TSP with many cities, the proposed algorithm outperforms the SA and the quantum annealing (QA) algorithm in terms of the optimization effect.

\section{Preliminaries}
\label{Preliminaries}
\subsection{Definitions of quantization}
\label{ch02-01}
Before illustrating the proposed algorithm, we establish the following definitions and assumptions.
\begin{definition}  
\label{def_01}
For $f \in \mathbf{R}$, we define the quantization of $f$ as follows:
\begin{equation}
f^Q \triangleq \frac{1}{Q_p} \lfloor Q_p \cdot  (f + 0.5 \cdot Q_p^{-1}) \rfloor
= \frac{1}{Q_p} \left( Q_p \cdot f + \varepsilon \right) = f + \varepsilon Q_p^{-1}, \quad f^Q \in \mathbf{Q}
\label{define_eq01}
\end{equation}
, where $\lfloor f \rfloor \in \mathbf{Z}$ denotes the floor function such that $\lfloor f \rfloor \leq f$ for all $f \in \mathbf{R}$, $Q_p \in \mathbf{Q}^+$ denotes the quantization parameter, and $\varepsilon \in \mathbf{R}$ represents the quantization error. 
\end{definition}
\begin{definition}  
\label{def_02}
We define the quantization parameter $Q_p \in \mathbf{Q}^+$ to be a monotone increasing function $Q_p : \mathbf{R}^{++} \mapsto \mathbf{Z}^{+}$ such that 
\begin{equation}
Q_p (t) = \eta \cdot b^{\bar{h}(t)} 
\label{def_eq02}    
\end{equation}
, where $\eta \in \mathbf{Q}^{++}$ denotes the fixed constant parameter of the quantization parameter, $b$ represents the base, and $\bar{h} :\mathbf{R}^{++} \mapsto \mathbf{Z}^+$ denotes the power function such that $\bar{h}(t) \uparrow \infty \; \text{ as } \; t \rightarrow \infty$.
\end{definition}
\begin{assumption}  
For a numerical sequence $\{ f(t)\}_{t=0}^{\infty}$ where each $f(t) \in \mathbf{R}^+ \; \forall t > 0$, suppose that $f(t)$ is defined on a dense topology space. Following the WNH and $\eqref{def_01}$, the quantization error $\varepsilon_t$ corresponding to $t > 0$ is i.i.d. white noise defined on the probability space $(\Omega, \mathcal{F}_t, \mathbb{P}_{\varepsilon})$. 
\label{assum_01}
\end{assumption}
Under Assumption $\ref{assum_01}$, we can regard the sequence of the $f^Q_t$ corresponding to $f(t)$ as a stochastic process $\{ f^Q_t \}_{t=0}^{\infty}$. 
As a next step, to analyze the properties of the stochastic process $\{ f^Q_t \}_{t=0}^{\infty}$, we calculate the mean and variance of the quantization error. 
\begin{theorem-opt}
\label{th_001}      
If the quantization error $\varepsilon_t \in \mathbf{R}^n$ satisfying the WNH, the mean and variance of the quantization error at $t > 0$ is 
\begin{equation}
\forall \varepsilon_t \in \mathbf{R}, \quad
\mathbb{E}_{\mathcal{F}_t} Q_p(t) \varepsilon_t = 0, \quad
\mathbb{E}_{\mathcal{F}_t} Q_p^{-2}(t) \varepsilon^2_t = Q_p^{-2}(t) \cdot \mathbb{E}_{\mathcal{F}_t} {\varepsilon}^2_t = \frac{1}{12 \cdot Q_p^2 (t)}. 
\label{eq_th01_1}
\end{equation}
\end{theorem-opt}
To discuss the main algorithm, we consider the optimization problem for an objective function $f$ such that
\begin{equation}
\text{minimize} \quad  f : \mathbf{R}^n \mapsto \mathbf{R}^+. 
\label{eq02_01}    
\end{equation}
In various combinatorial optimization problems, we deal with an actual input represented as $x^r \in [0, 1]^m$. Thus, we suppose that there exists a proper transformation from a binary input to a proper real vector space such that $\mathcal{T} : [0,1]^m \rightarrow \mathcal{X} \subseteq \mathbf{R}^n.$, where $\mathcal{X}$ represents the virtual domain of the objective function $f$.
Under the transformation assumption, we assume that $f \in C^{\infty}$ fulfills the Lipschitz continuity as follows:
\begin{assumption}  
\label{assum02}
For $x_t \in B^o (x^*, \rho)$, there exists a positive value $L$ w.r.t. a scalar field $f(x) : \mathbf{R}^n \rightarrow \mathbf{R}$ such that
\begin{equation}
\| f(x_t) - f(x^*) \| \leq L \| x_t - x^* \|, \quad \forall t > t_0, 
\label{eq_as02_02}    
\end{equation}
where $B^o (x^* , \rho)$ denotes an open ball $B^o (x^* , \rho) = \{ x | \| x - x^* \| < \rho \}$ for all $\rho \in \mathbf{R}^{++}$, and $x^* \in \mathbf{R}^n$ denotes the globally optimal point.
\end{assumption}

\subsection{Primitive algorithm}
\label{ch02-02}
\begin{algorithm2e}[bt]
\SetAlgoLined
\DontPrintSemicolon
\caption{Blind Random Search (BRS) with the proposed quantization scheme}\label{alg-1}
\begin{multicols}{2}
    \KwInput{Objective function $f(x) \in \mathbf{R}^{+}$}
    \KwOutput{$x_{opt}, \; f(x_{opt})$}
    \KwData{$x \in \mathbf{R}^n$}
    {\bfseries Initialization} \; 
     $t \leftarrow 0$ and $\bar{h}(0) \leftarrow 0$ \;
     Set initial candidate $x_0$ and $x_{opt} \leftarrow x_0$\;
     Compute the initial objective function $f(x_0)$ \;
     Set $b=2$ and $\eta= b^{-\lfloor \log_b (f(x_0) + 1)\rfloor}, \; Q_p \leftarrow \eta$\;
     $f^Q_{opt} \leftarrow \frac{1}{Q_p} \lfloor Q_p \cdot  (f + 0.5 \cdot Q_p^{-1}) \rfloor$\;
    \While{Stopping condition is satisfied}{
         Set $t \leftarrow t+1$ \;
         Select $x_t$ randomly and compute $f(x_t)$\;
         $f^Q \leftarrow \frac{1}{Q_p} \lfloor Q_p \cdot  (f + 0.5 \cdot Q_p^{-1}) \rfloor$\;
        \If {$f^Q \leq f^Q_{opt}$ }{     
             $x_{opt} \leftarrow x_t$ \;
             $\bar{h}(t) \leftarrow \bar{h}(t) + 1, \; Q_p \leftarrow \eta \cdot b^{\bar{h}(t)}$ 
             $f^Q_{opt} \leftarrow \frac{1}{Q_p} \lfloor Q_p \cdot  (f + 0.5 \cdot Q_p^{-1}) \rfloor$ \;
        }    
    }
\end{multicols}
\end{algorithm2e}

As the most elementary implementation, we apply the proposed quantization scheme to the blind random search(BRS) algorithm.

First, as shown in Algorithm $\ref{alg-1}$, we randomly select an input point $x_{t}$ and quantize the value of the objective function $f(x_{t})$ such that $f^Q(x_{t})$ with the quantization parameter $Q_p(t-1)$.  
Comparing both quantization values $f^Q(\bar{x}_{t-1})$ and $f^Q(x_{t})$, if $f^Q(\bar{x}_{t-1})$ is greater than or \textbf{equal} to $f^Q(x_t)$, we set ${x}_t$ to be the optimal value and substitute $\bar{x}_t$ to the $x_t$. 
Following this procedure, we update the quantization parameter as $Q_p(t-1)$ by increasing the power function $\bar{h}(t)$ defined in $\eqref{def_02}$. 
We denote it as the re-quantization process. 
Because we update the quantization parameter, the quantization value of $f^Q(x_t)$ is re-quantized with $Q_p(t)$. 
Consecutively, we select another input point as a part of the BRS. 

Furthermore, we propose a simple initialization of the quantization parameter to implement the BRS using the proposed scheme.  
We want the transition probability of the initial state $\mathbb{P}(x_1 | x_0)$ to be a high such as $\mathbb{P}(x_1 | x_0)=1$. 
Therefore, the quantization of the other objective function value $f^Q(x_1) \forall x_1 \ne x_0 $ should be lower than the quantization of the initial objective function. 
For this purpose, we set the initial parameter of the quantization parameter $\eta$, as represented in the following theorem:
\begin{theorem-opt}
\label{th02}            
Suppose that the initial value of a given objective function $f(x_0) \in \mathbf{R}$ is $\sup_{x \in \mathbf{R}} f(x) $. 
The transition probability to the next step, $\mathbb{P}(x_1|x_0)$, yielded by the proposed algorithm, is one when the initial parameter $\eta \in \mathbf{Q}^{+}$ satisfies the following equation:
\begin{equation}
\label{eq_th02_01}
    \eta = b^{-\lfloor log_b (f(x_0) + 1) \rfloor}
\end{equation}
, where $b$ represents a base in Definition $\ref{def_02}$ for $Q_p(t)$.
\end{theorem-opt}
\section{Analysis of the proposed algorithm}
\label{Analysis}
\subsection{Fundamental dynamics of the proposed algorithm}
Let a subset of the virtual domain $\mathcal{X} \subseteq \mathbf{R}^n$ such that $L^Q(t) \triangleq \{x_t | f(x) -  f^Q(\bar{x}_{t}) \leq 0, Q_p(t) \}$. 
Following the procedure in Algorithm 1 and the definition of the subset $L^Q(t)$, we note that the proposed algorithm can yield the following containment relationship between subsets:
\begin{equation}
\label{eq31_01}
\exists t > t_0, \; L^Q(t) \supseteq L^Q(t+1) \cdots \supseteq L^Q(t+k)
\end{equation}
The above equation can lead to the measure of $L^Q(t)$ being proportional to $Q_p^{-1}(t)$ following the Lipschitz continuous property represented in Assumption $\ref{assum02}$.
In addition, as $Q_p^{-1}(t)$ decreases monotonically by Definition $\ref{def_02}$, we can obtain  the following inequalities about the measure of each subset: 
\begin{equation}
\exists t > t_0, \; m(L^Q(t)) \geq m(L^Q(t+1)) \cdots \geq m(L^Q(t+k)). 
\label{eq31_02}
\end{equation}
Suppose that there exists a unique optimizer $x^*$ such that $\forall x \in \mathcal{X}, \;f(x^*) \leq f(x) \leq f^Q(x)$.  
If $\eqref{eq31_02}$ results in $\lim_{t \uparrow \infty} m(L^Q(t+k))=0$, $f^Q(x) \rightarrow f(x)$. Accordingly, following the above assumption of a unique optimizer, we can obtain $f^Q(x) \rightarrow f(x^*)$ intuitively. 

To prove the above consideration, we establish the following assumption.
\begin{assumption}
\label{assum_03}
The power summation to the base $b^{\bar{h}(t+k)}$ is bounded such that
\begin{equation}
\label{as03-eq01}
\lim_{k \rightarrow \infty} \sum_{k=0}^n b^{-\bar{h}(t+k)} = \bar{b}(t) < \infty, \quad \bar{b}(t) \downarrow 0 \text{ as } t \uparrow \infty
\end{equation}
\end{assumption}
Under Assumption $\ref{assum_03}$, we can  establish the following theorem
\begin{theorem-opt}
\label{th_003}
For a large $k > n_0$, if the proposed algorithm provides a sufficiently finite resolution for $f^Q$ such that
\begin{equation}
f^Q(x_{t+k}) - f^Q(x_{t+k+1}) = Q_p(t+k)^{-1}
\end{equation}
, for all $x_t \in \mathbf{R}^n$ and $t > 0$, there exists $n < n_0$ satisfying the following
\begin{equation}
\label{th03_eq00}
\| f(x_{t+n}) - f(x_{t+n+1}) \| \geq \| f(x_{t+k}) - f(x^*) \|.
\end{equation}
\end{theorem-opt}
For the stochastic analysis of the proposed algorithm, we can obtain the following lemma associated with the difference of quantization errors to the quantized objective functions.  
\begin{lemma-opt}
\label{Lm_04}
Suppose that there exist two equal quantized objective functions for two distinguished inputs $x_t, x_{t+1} \in \mathbf{R}^n$ such that $f^Q(x_t) = f^Q(x_{t+1})$.
Under this condition, the quantization error $\bar{\varepsilon}_t Q_p^{-1}(t)$ of $f^Q(x_t) - f^Q(x_{t+1})$ is evaluated as follows:
\begin{equation}
\label{Lm_eq00}
\bar{\varepsilon}_{t} Q_p^{-1} = -(\varepsilon_{t+1} - \varepsilon_t) \cdot (x_{t + 1} - x_{t}) \cdot v_t \cdot \tilde{Q}_p^{-1}, \quad \bar{\varepsilon}_t \triangleq \varepsilon_{t+1} - \varepsilon_t
\end{equation}
, where $v_t$ represents a normalized vector defined as $v_t = -\frac{x_{t+1} - x_t}{\| x_{t+1} - x_t \|}$ and $\tilde{Q}_p(t)$ denotes a scaled quantization parameter to $Q_p(t)$ with a constant value $C \in \mathbf{R}^+$ such that
\begin{equation}
\tilde{Q}_p^{-1}(t) = C \cdot b^{\bar{h}(t)}.    
\end{equation}
\end{lemma-opt}
With the above theorem and lemma, we can establish the stochastic differential equation(SDE) for the proposed algorithm as follows:
\begin{theorem-opt}
\label{th_005}
For a given objective function $f(x_t) \in \mathbf{R}$, suppose that there exist the quantized objective functions $f^Q(x_t), \; f^Q(x_{t+1})$ at a current state $x_t$ and the following state $x_{t+1}$ such that $f^Q(x_t) \geq f^Q(x_{t+1})$, for all $x_{t+1} \neq x_t$; we can obtain the differential equation of the state transition as follows:
\begin{equation}
\label{eq_th05_01}
dX_t = - \nabla_x f(X_t) dt + \sqrt{C_q} \cdot Q_p^{-1}(t) dW_t
\end{equation}
where $W_t \in \mathbf{R}^n$ represents a standard Wiener process, which has a zero mean and variance with one, $X_t \in \mathbf{R}^n$ denotes a random variable corresponding to $x_t \in \mathbf{R}^n$, and $C_q \in \mathbf{R}$ is a constant value.
\end{theorem-opt}

\subsection{Weak convergence of the proposed algorithm}
Equation $\eqref{eq_th05_01}$ is the typical Langevine SDE, so we can expect that the transition probability yielded by the proposed algorithm follows Gibb's distribution based on a Gaussian function.
In addition, we note that the proposed algorithm exhibits the hill-climbing effect resulting from the Wiener process $dW_t$; thus, the proposed algorithm is robust to local minima \cite{Lundy_1986, Locatelli_1996}.
However, an asymptotic analysis of the Hilbert space always represents the possibilities of divergence in an optimization algorithm with the hill-climbing property.
Therefore, we demonstrate a global optimization of the algorithm, including the hill-climbing, which is robust to local minima so that we prove the convergence of the transition probability yielded by the proposed algorithm to the global optimum. 
This convergence is known as a weak convergence. 
In particular, based on the Laplace theorem, the proof of weak convergence to the transition probability represented by Gibbs's distribution is clear .
In the proposed algorithm, as shown in $\eqref{eq_th05_01}$, the variance of the transition probability is in proportional to the inverse of the quantization parameter $Q_p^{-1}(t)$. 
The inverse of the quantization parameter is a monotone decreasing function to time $t$, as represented in Definition $\ref{def_02}$, and the limit of the summation to time is finite, as shown in Assumption $\ref{assum_03}$.
Consequently, we can expect that the proposed algorithm fulfills Laplace's theorem(\cite{Laplace_1980, Geman-1986, Hwang-1987, Locatelli_1996}), and we can prove the weak convergence as follows:
\begin{theorem-opt}
\label{th_006}
If the dynamics of the state transition by the proposed algorithm follow $\eqref{eq_th05_01}$, the state $x_t$ weakly converges to the global minimum when the quantization parameter decreases to the following schedule:
\begin{equation}
\label{eq_th_06}    
\inf_{t \geq 0} Q_p^{-1}(t) = \frac{C_o}{\log (t + 2)}, \quad C_o \in \mathbf{R}^+,\; C_o \gg 0
\end{equation}
\end{theorem-opt}
With the assumption of an objective function's Lipschitz continuous property, we can prove Theorem 6 without any convex assumptions.
Another property shown by Theorem 6 is that the proposed primitive algorithm contains more strong convergence conditions than Theorem 6 represents. 
In addition, because $Q_p^{-1}$ is not a rational number, the implementation of the proposed algorithm may be elusive. 
Therefore, by letting $\eqref{eq_th_06}$ be an upper bound and another equation be a lower bound, we can set $\bar{h}(t)$ as the following theorem for the global convergence and implementation of the proposed algorithm.
\begin{theorem-opt}
\label{th_007}
Suppose that there exists an integer valued annealing schedule $\sigma(t) \in \mathbf{Z}^+$ such that $\sigma(t) \geq \inf \sigma(t) \triangleq c/\log(t+2)$.  If the power function $\bar{h}(t)$ of the quantization parameter $Q_p^{-1}(t)$ fulfills the following condition, the proposed algorithm weakly converges to the global optimum.
\begin{equation}
    \log_b \left(C_0 \cdot b^{-\frac{2 \beta}{t+2}} \cdot \inf \sigma(t) \right) \leq \bar{h}(t) \leq \log_b \left( C_1 \log(t+2) \right)
\end{equation}
, where $C_0 \equiv \eta \sqrt{C_q}$ and $C_1 \equiv \sqrt{C_q}\eta/C$.
\end{theorem-opt}
Theorem $\ref{th_007}$ illustrates that if the algorithm controls the power function $\bar{h}(t)$  for the quantization under the condition in Theorem $\ref{th_007}$, we can find the global optimum with a weak convergence property.
\section{Simulation Results}
\begin{table}
\scriptsize
\caption{Simulation Results to TSP for 100 cities}\label{result-1}
\centering{
\begin{tabular}{lccc}
\toprule
Criterion & Simulated Annealing & Quantum Annealing & Proposed Algorithm\\
\midrule
Average Minimum Cost                     &  1729.50   & 1721.07     & 1648.26  \\
Improvement Ratio to the Initial setting &  19.90\%   & 20.29\%     & 23.67 \% \\ 
\bottomrule                                   
\end{tabular}
}
\end{table}

\begin{table}
\scriptsize
\caption{Simulation Results to TSP beyond 100 cities}\label{result-2}
\centering{
\begin{tabular}{lccccc}
\toprule
Number of Cities & Nearest Neighbor(Initial)& Simulated Annealing & Quantum Annealing   & Proposed Algorithm & Improve Ratio \\
\midrule
100    & 2159.27    &1729.50 &  1721.07 & 1648.26 & 23.67        \\
125    & 2297.86    &2027.52 &	2028.2  & 1923.65 &	16.28        \\
150    & 2497.65    &2255.15 &	2252.82 & 2032.21 &	18.63        \\
175    & 2380.52    &2380.52 &	2380.29	& 2147.17 &	9.80         \\
200    & 2769.73    &2769.34 &	2769.42 & 2366.72 &	14.55        \\
\bottomrule
\end{tabular}
}
\end{table}

To verify the optimization performance of the proposed algorithm for combinatorial optimization problems, including NP-hard problems, we perform the TSP simulation of 100 cities located in a 2-dimensional squared space with the range $[0, 200]$.  
We use the OPT-2 algorithm, which is the city selection method in the TSP for cost evaluation as presented by \cite{Johnson_1997}.
The OPT-2 algorithm is one of the transform functions for real binary input space to a virtual real vector space such as $\mathcal{T}:\{0,1\}^m \rightarrow \mathbf{R}^n$, where $m$ represents the number of cities minus 1, $n$ represents a virtual dimension of the virtual space.
Using such a transformation, we can assume that an objective function fulfills Lipschitz continuity in that the OPT-2 algorithm changes the location of only two cities\cite{Geman-1986}.
In all attempts, we use a fixed location of cities to guarantee the generality of the simulation.
Moreover, to guarantee objective optimization performance for all algorithms in the simulation,  we set an initial route for each city in the TSP using the nearest neighbor algorithm, and we set the initial route as a start Hamiltonian $H_0$ for QA.
The simulation results in Table $\ref{result-1}$ show that the average optimization performance of the proposed algorithm is superior to that of SA and QA.  

Furthermore, we evaluate the optimization performance of the algorithms in solving a TSP involving more than 100 cities.
Generally, the difficulties in a TSP involving more than 100 cities increase dramatically.  
For instance, the possible number of routes in the TSP from 100 to 110 cities increase approximately $10^{20}$ times (from $9.33 \times 10^{157}$ to $1.58 \times 10^{178}$). 
Such increasing difficulties in the TSP lead to high computational costs and optimization failure.
The simulation results in Table $\ref{result-2}$ show that the proposed algorithm can find a feasible solution even when the number of cities is 200, whereas other algorithms are unable to outperform the nearest neighbor method in finding a better solution.

\section{Conclusion}
We present a quantization-based optimization scheme with an increase in the quantization resolution to optimize an objective function globally.
Through stochastic analysis, particularly the SDE, the dynamics of the proposed algorithm, are described.  
Using the SDE and feasible assumptions, we present the analysis of the weak convergence of the proposed algorithm, enabling global optimization.
The proposed algorithm is based on the mathematical feature of quantization error, whereas other heuristic algorithms simulate natural phenomena.
Therefore, we expect to develop an alternative global optimization methodology by numerical analysis based on number theory. 
In future work, we will research an effective iterative difference learning equation based on a quantized optimization scheme for a continuous function defined on the differential manifold.

\newpage



\newpage
\appendix

\newtheorem{theorem-ack}{Theorem}
\newtheorem{lemma-ack}[theorem-ack]{Lemma}
\newtheorem{lemma-aux}{Lemma : Auxiliary}
\section{Acknowledgement}
This work was supported by Institute for Information and communications Technology Promotion(IITP) grant funded by the Korea government(MSIP) (2021-0- 00766, Development of Integrated Development Framework that supports Automatic Neural Network Generation and Deployment optimized for Runtime Environment)
\section{Introduction}
We set notations, proof of lemmas and theorems and more detailed information about the simulation in the manuscript to the following sections.
\section{Notations}
\begin{itemize}
    \item $\mathbf{R}^n$  ~ The n-dimensional space with real numbers
    \item $\mathbf{R}$    ~ $\mathbf{R}^n \vert_{n=1}$
    \item $\mathbf{R}[\alpha, \beta]$ ~ $\{ x \in \mathbf{R} | \alpha \leq  x \leq \beta, \; \alpha, \beta \in \mathbf{R} \}$
    \item $\mathbf{R}(\alpha, \beta]$ ~ $\{ x \in \mathbf{R} | \alpha <  x \leq \beta, \; \alpha, \beta \in \mathbf{R} \}$
    \item $\mathbf{R}[\alpha, \beta)$ ~ $\{ x \in \mathbf{R} | \alpha \leq  x < \beta, \; \alpha, \beta \in \mathbf{R} \}$
    \item $\mathbf{R}(\alpha, \beta)$ ~ $\{ x \in \mathbf{R} | \alpha <  x < \beta, \; \alpha, \beta \in \mathbf{R} \}$
    \item $\mathbf{Q}^n$  ~ The n-dimensional space with rational numbers
    \item $\mathbf{Q}$    ~ $\mathbf{Q}^n \vert_{n=1}$
    \item $\mathbf{Z}$    ~ The 1-dimensional space with integers. 
    \item $\mathbf{N}$    ~ The 1-dimensional space with natural numbers.  
    \item $\mathbf{R}^+$  ~ $\{ x \vert x \geq 0, \;  x \in \mathbf{R}\}$
    \item $\mathbf{R}^{++}$ ~ $\{ x \vert x > 0, \;  x \in \mathbf{R}\}$
    \item $\mathbf{Q}^+$  ~ $\{ x \vert x \geq 0, \;  x \in \mathbf{Q}\}$
    \item $\mathbf{Q}^{++}$ ~ $\{ x \vert x > 0, \;  x \in \mathbf{Q}\}$
    \item $\mathbf{Z}^+$  ~ $\{ x \vert x \geq 0, \;  x \in \mathbf{Z}\}$
    \item $\mathbf{Z}^{++}$ ~ $\{ x \vert x > 0, \;  x \in \mathbf{Z}\}$,  $\mathbf{Z}^{++}$ is equal to $\mathbf{N}$.
    \item $\lfloor x \rfloor$ ~ $\max \{ y \in \mathbf{Z}| y \leq x, \forall x \in \mathbf{R} \}$
    \item $\lceil x \rceil$ ~ $\min \{ y \in \mathbf{Z}| y \geq x, \forall x \in \mathbf{R} \}$    
\end{itemize}

\section{Auxiliary Lemma }
We use the following lemma to prove the theorems represented in the next chapters. 
\begin{lemma-aux}
\label{eq01:lemma}
For all $x \in \mathbf{R}$,
\begin{equation}
    ( 1 - x) \leq \exp (-x).
\end{equation}
\end{lemma-aux}
\begin{proof}
By the definition of the exponent, we write the exponential function as the following fundamental series : 
\begin{equation}
    \exp (-x) = \sum_{n=0}^{\infty} \frac{1}{n!} (-1)^n x^n  = \sum_{k=0}^{\infty} \left( \frac{1}{2k!} x^{2k} - \frac{1}{(2k+1)!} x^{2k+1}\right).
\end{equation}
Let $u_k$ as follows:
\begin{equation}
    u_k = \frac{1}{2k!} x^{2k} \left( 1 - \frac{1}{2k+1} x \right)
\end{equation}
then we can rewrite the series of exponents such that 
\begin{equation}
    \exp(-x) = u_0 + \sum_{k=1}^{\infty} u_k.
\end{equation}
For all $k > 0$, since each $u_k$ is positive, we have
\begin{equation}
    1 - x = u_0  \leq u_0 + \sum_{k=0}^{\infty} u_k.
\end{equation}
Alternatively, we can prove the lemma with differentiation.
Let $g(x) = (1 - x) - \exp(-x)$. Differentiating $g(x)$ to $x$, we get
\begin{equation}
    \frac{d g}{dx}(x) = -1 + exp(-x), \; \frac{d^2 g}{dx^2} = - \exp(-x)
\end{equation}
We note that $g(x)$ is a concave function from the fact that $\frac{d^2 g}{dx^2} < 0, \;  \forall x \in \mathbf{R}$. In addition, the maximum of $g(x)$ is zero at $x=0$ from which $\frac{d g}{dx}(x) = -1 + exp(-x) = 0$.
Therefore, $g(x) \leq 0$, so that it fulfills the Lemma. 
\end{proof}
\section{Proofs of Theorems in Section 2}
\subsection{Proof of theorem 1}
\begin{theorem-ack}
If the quantization error $\varepsilon_t \in \mathbf{R}^n$ satisfying the WNH, the mean and variance of the quantization error at $t > 0$ is 
\begin{equation}
\forall \varepsilon_t \in \mathbf{R}, \quad
\mathbb{E}_{\mathcal{F}_t} Q_p(t) \varepsilon_t = 0, \quad
\mathbb{E}_{\mathcal{F}_t} Q_p^{-2}(t) \varepsilon^2_t = Q_p^{-2}(t) \cdot \mathbb{E}_{\mathcal{F}_t} {\varepsilon}^2_t = \frac{1}{12 \cdot Q_p^2 (t)}. 
\end{equation}
\end{theorem-ack}
\begin{proof}
The theorem is explicit according to the WNH. 
Let $\Delta$ be the brief notation of $\varepsilon_t Q_p^{-1}(t)$. 
According to \citet{Jimnez_2007}, $\varepsilon_t$ is uniformly distributed in $[-Q_p^{-1}(t), Q_p^{-1}(t))$ under the WNH and Definition $\ref{def_01}$.
Therefore, we can obtain the expectation value of $\Delta = \varepsilon_t Q_p^{-1}(t)$ as follows:
\begin{equation}
    \mathbb{E}_{\mathcal{F}_t} \Delta 
    = \int_{-Q_p^{-1}(t)/2}^{Q_p^{-1}(t)/2} \Delta \mathbb{P}_{\varepsilon} d\Delta 
    = \frac{1}{Q_p^{-1}(t)} \cdot \int_{-Q_p^{-1}(t)/2}^{Q_p^{-1}(t)/2} \Delta  d\Delta 
    = \frac{1}{2 Q_p^{-1}(t)} \left( \frac{Q_p^{-1}(t)^2}{2^2 } - \frac{Q_p^{-1}(t)^2}{(-2)^2} \right) = 0.
\end{equation}

In a similar way, we can obtain the variance such that
\begin{equation}
    \mathbb{E}_{\mathcal{F}_t} \Delta^2 
    = \frac{1}{Q_p^{-1}(t)} \int_{-Q_p^{-1}(t)/2}^{Q_p^{-1}(t)/2} \Delta^2 d\Delta 
    = \frac{1}{Q_p^{-1}(t)} \cdot \frac{1}{3} \left( \frac{Q_p^{-1}(t)^3}{8 } - \frac{-Q_p^{-1}(t)^3}{8 } \right) 
    = \frac{1}{12 \cdot Q_p^2(t)}  
\end{equation}
From the WNH, the square of $\varepsilon_t$ is one, so that we obtain the result of the theorem.
\end{proof}

\subsection{Proof of theorem 2}
\begin{theorem-ack}
Suppose that the initial value of a given objective function $f(x_0) \in \mathbf{R}$ is $\sup_{x \in \mathbf{R}} f(x) $. 
The transition probability to the next step, $\mathbb{P}(x_1|x_0)$, yielded by the proposed algorithm, is one when the initial parameter $\eta \in \mathbf{Q}^{+}$ satisfies the following equation:
\begin{equation}
    \eta = b^{-\lfloor log_b (f(x_0) + 1) \rfloor}
\label{eq_th3_2}
\end{equation}
, where $b$ represents a base in Definition $\ref{def_02}$ for $Q_p(t)$.
\label{th3_2}
\end{theorem-ack}

\begin{proof}
By the assumption in Theorem $\ref{th02}$, we can establish the following inequality for all $x_1 \neq x_0$
\begin{equation}
    f(x_0) + Q_p^{-1} (0) \geq f(x_1) + Q_p^{-1}(1).
    \label{th3_2pf01}
\end{equation}
By the definition of the quantization parameter $Q_p$, $Q_p(0) = \eta b^{0} = \eta$ and $Q_p(1) = \eta b^{-1}$; thus,
\begin{equation}
    f(x_0) + \eta^{-1} \geq f(x_1) + \eta^{-1} b \implies  f(x_0) - f(x_1) \geq \eta^{-1} (b - 1). 
    \label{th3_2pf02}
\end{equation}
Suppose that $\eta$ is the power of $b$, i.e., $\eta = b^k$, where $k \in \mathbf{Z}^{+}$, 
we can get
\begin{equation}
\begin{aligned}
    f(x_0) - f(x_1) \geq b^{-k} (b - 1) 
    &\implies \frac{f(x_0) - f(x)}{b - 1} \geq b^{-k} \\
    \implies -\log_b \frac{f(x_0) - f(x_1)}{b - 1} \leq k 
    &\implies k \geq \log_b (b -1) - \log_b ( f(x_0) - f(x) ). 
    \label{th3_2pf03}
\end{aligned}
\end{equation}
Since $\log_b (b-1) \geq 0$ and $\log_b (f(x_0) - f(x_1)) \geq  \log_b f(x_0)$ for all $x_1 \neq x_0$, we obtain
\begin{equation}
    k \geq \log_b (b -1) - \log_b ( f(x_0) - f(x)) > -1 - \log_b f(x_0) \geq - \lfloor 1 + \log_b f(x_0) \rfloor
    \label{th3_2pf04}
\end{equation}

Therefore, if $f(x_0) \in \mathbf{R}$ is $\sup_{x \in \mathbf{R}} f(x)$, the initial transition probability is one, andwe can set establish initial value of quantization parameter $\eta = Q_p(0)$ to be
\begin{equation}
    \eta = b^{-\lfloor log_b (f(x_0) + 1 \rfloor}
\end{equation}
\end{proof}

\section{Proof of Lemma and Theorems in Section 3}
\subsection{Proof of theorem 3}
\begin{theorem-ack}
For a large $k > n_0$, if the proposed algorithm provides a sufficiently finite resolution for $f^Q$ such that
\begin{equation}
f^Q(x_{t+k}) - f^Q(x_{t+k+1}) = Q_p(t+k)^{-1}
\end{equation}
, for all $x_t \in \mathbf{R}^n$ and $t > 0$, there exists $n < n_0$ satisfying the following
\begin{equation}
\| f(x_{t+n}) - f(x_{t+n+1}) \| \geq \| f(x_{t+k}) - f(x^*) \|.
\end{equation}
\end{theorem-ack}
\begin{proof}           
Assume that $f^Q(x^*) = f(x^*)$, and $f^Q(x) \neq f^Q(y)$ for all $x, y \in \mathbf{R}^n$ and $x \neq y$.
From the definition of the algorithm, we let $Q_p(\tau)^{-1}$ denote the infimum of the difference between $f^Q(x)$ and $f^Q(y)$ when $f^Q(x) \neq f^Q(y)$; thus, we can obtain
\begin{equation}
f^Q(x_s) - f^Q(x_{s+1})
\geq Q_p(s)^{-1}
= \eta^{-1} \cdot b^{-\bar{h}(s)}
, \quad \forall b \in \mathbf{Z}(1, \infty)
\label{pf-th2-eq01}
\end{equation}
, where $s \in \mathbf{Z}^+$.
By the assumption in Theorem $\ref{th_003}$, the difference $f^Q(x_\tau) - f^Q(x_{\tau+1})$ is equal to the each quantization step i.e. $f^Q(x_\tau) - f^Q(x_{\tau+1}) = \eta^{-1} \cdot b^{-\bar{h}(\tau)}$, for an positive real integer $\tau > s$.   
Accordingly, $\eqref{pf-th2-eq01}$ leads
\begin{equation}
\begin{aligned}
f^Q(x_\tau) - f^Q(x^*)
&= f^Q(x_\tau) - f^Q(x_{\tau+1}) + f^Q(x_{\tau+1}) - \cdots - f^Q(x_{\tau+n}) + f^Q(x_{\tau+n}) - f^Q(x^*) \\
&= \eta^{-1} \sum_{k=0}^{n-1}  b^{-\bar{h}(\tau+k)} + f^Q(x_{\tau+n}) - f^Q(x^*).
\end{aligned}
\end{equation}

If we can find the optimal point at the step $\tau+n$, we can obtain the supremum of the bound to the difference $f^Q(x_{\tau+n}) - f^Q(x^*)$ as follows:
\begin{equation}
\begin{aligned}
\sup \inf_{x_{\tau+n}} \| f^Q(x_{\tau+n}) - f^Q(x^*) \|
&= \sup \inf_{x_{\tau+n}} \| f^Q(x_{\tau+n}) - f(x^*) \| \\
&= \sup \inf_{x_{\tau+n}} \| f(x^*) + \varepsilon Q_p^{-\bar{h}(\tau+n)} - f(x^*) \| \\
&= Q_p^{-\bar{h}(\tau+n)} = \eta^{-1} \cdot b^{-\bar{h}(\tau+n)}.
\end{aligned}
\end{equation}
Thus, we can obtain
\begin{equation}
\begin{aligned}
f^Q(x_\tau) - f^Q(x^*)
&\leq \eta^{-1} \sum_{k=0}^{n-1}  b^{-\bar{h}(\tau+k)} + \eta^{-1} \cdot b^{-\bar{h}(\tau+n)} \\
&= \eta^{-1} \sum_{k=0}^{n}  b^{-\bar{h}(\tau+k)} 
< \eta^{-1} \sum_{k=0}^{\infty}  b^{-\bar{h}(\tau+k)} 
= \eta^{-1} \cdot \bar{b}(\tau).  
\end{aligned}
\end{equation}
Since the $\bar{b}(t)$ is a monotone decreasing function with respect to $t$, there exists $\delta > 0$ such that $\delta > \bar{b}(\tau)$.
Therefore, there exists $s > \tau$ such that
\begin{equation}
  f^Q(x_s) - f^Q(x_{s+1}) \geq \eta^{-1} \cdot b^{-\bar{h}(s)} \geq \eta^{-1} \cdot \delta > \eta^{-1} \cdot \bar{b}(\tau) > f^Q(x_\tau) - f^Q(x^*).
\end{equation}
\end{proof}

\subsection{Proof of lemma 4 }
\begin{lemma-ack}
Suppose that there exist two equal quantized objective functions for two distinguished inputs $x_t, x_{t+1} \in \mathbf{R}^n$ such that $f^Q(x_t) = f^Q(x_{t+1})$.
Under this condition, the quantization error $\bar{\varepsilon}_t Q_p^{-1}(t)$ of $f^Q(x_t) - f^Q(x_{t+1})$ is evaluated as follows:
\begin{equation}
\bar{\varepsilon}_{t} Q_p^{-1} = -(\varepsilon_{t+1} - \varepsilon_t) \cdot (x_{t + 1} - x_{t}) \cdot v_t \cdot \tilde{Q}_p^{-1}, \quad \bar{\varepsilon}_t \triangleq \varepsilon_{t+1} - \varepsilon_t
\end{equation}
, where $v_t$ represents a normalized vector defined as $v_t = -\frac{x_{t+1} - x_t}{\| x_{t+1} - x_t \|}$ and $\tilde{Q}_p(t)$ denotes a scaled quantization parameter to $Q_p(t)$ with a constant value $C \in \mathbf{R}^+$ such that
\begin{equation}
\tilde{Q}_p^{-1}(t) = C \cdot b^{\bar{h}(t)}.    
\end{equation}
\end{lemma-ack}
\begin{figure}
    \centering
    \resizebox{\textwidth}{!}{
    \includegraphics[]{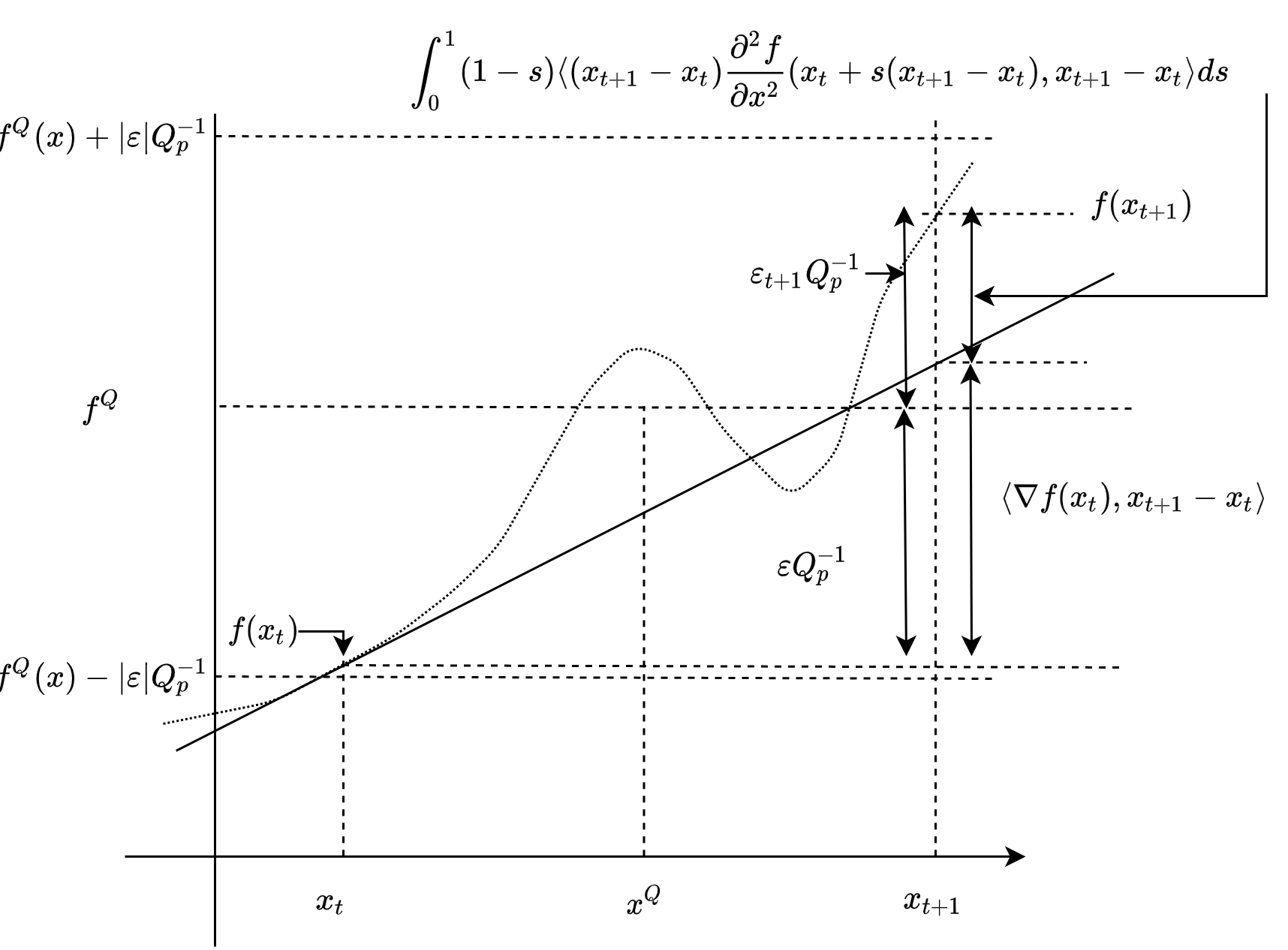}
    }
    \caption{Conceptual diagram for Lemma $\ref{Lm_04}$. In an equal quantization level, we can dismiss correct value of $f(x)$ since $f^Q(x) = f^Q(y), \forall x \neq y$.Thus, we can let $f(x)$ as a simple low-order function such as the first-order function, within an equal quantization level instead of the correct $f(x)$}
    \label{fig_lm04_01}
\end{figure}
\begin{proof}
According to the assumption, we can consider the case represented in the figure $\ref{fig_lm04_01}$ as follows:
\begin{equation}
\begin{aligned}
&0 = f^Q(x_{t+1}) - f^Q(x_{t}) = f(x_{t+1}) - f(x_t) + (\varepsilon_{t+1} - \varepsilon_t)Q_p^{-1} \\
&\implies
f(x_{t+1}) - f(x_t) = - (\varepsilon_{t+1} - \varepsilon_t)Q_p^{-1}.
\end{aligned}
\label{a2_eq01}
\end{equation}
Furthermore, considering the line across the points $(x_t, f(x_t))$ and $(x_{t+1}, f(x_{t+1}))$, we get the following equation:
\begin{equation}
\bar{f}'(x) = -\frac{f(x_{t+1}) - f(x_t)}{\| x_{t+1} - x_t \|} v_t \cdot (x - x_t) + f(x_t), \quad \because v_t = - \frac{x_{t+1} - x_t}{\| x_{t+1} - x_t \|}.
\label{a2_eq02}
\end{equation}
By the definition of the quantization parameter, we note that $Q_p (t) = \eta \cdot b^{\bar{h}(t)}$.
In addition, Theorem $\ref{th3_2}$ illustrates that $\eta = b^{-\lfloor \log_b (f(x_0) + 1) \rfloor}$.
Without losing generality, we can establish $\eta$ as follows:
\begin{equation}
  \eta = b^{-\lfloor \log_b (f(x_0) - f(x^*)) + 1 \rfloor}, \quad \forall x \in \mathbf{R}^n, f(x^*) < f(x).
\end{equation}
Practically, we cannot know the optimal point correctly in most optimization problems, so the above definition for $\eta$ is an ideal and theoretical case.
Therefore, we can let $Q_p^{-1}(t) = \eta^{-1} \cdot b^{-\bar{h}(t)}$. 
Under the condition of $Q_p^{-1}(t)$, we can get
\begin{equation}
Q_p^{-1}
= b^{\log_b (f(x_0) - f(x^*)) + \epsilon} \cdot b^{\bar{h}(t)}
= (f(x_0) - f(x^*)) \cdot b^{\bar{h}(t) + \epsilon}
\label{a3_eq00}
\end{equation}
, where $\epsilon$ denotes an error led by the floor operation.
Spanning $f(x_0) - f(x^*)$ for a finite value $n > n_0$, we get
\begin{equation}
\begin{aligned}
f(x_0) - f(x^*)
&= f(x_0) - f(x_1) + f(x_1) - f(x_2) \cdots + f(x_{t+n}) - f(x^*) \\
&\leq \| f(x_0) - f(x_1) \| + \| f(x_1) - f(x_2) \| \cdots + \| f(x_{t+n-1}) - f(x^*) \|
\end{aligned}
\end{equation}
Holding Theorem $\ref{th_003}$, we can rewrite the final term of the right side such that
\begin{equation}
\begin{aligned}
f(x_0) - f(x^*)
&\leq \| f(x_0) - f(x_1) \| + \| f(x_1) - f(x_2) \| \cdots + \| f(x_{t+n}) - f(x_{t+n-1}) \|\\
&\leq n \cdot \| f(x_{\tau+1}) - f(x_{\tau}) \|
\end{aligned}
\end{equation}
, where $\tau \in \mathbf{Z}^+$ is defined as
\begin{equation}
  \forall t > t_0, \exists \tau \in \mathbf{Z}[t_0, t] \; \text{such that} \; \| f(x_{\tau+1}) - f(x_{\tau}) \| > \| f(x_{t+1}) - f(x_{t}) \|.
\end{equation}
By Lipschitz continuity, we note $\| f(x_{\tau+1}) - f(x_{\tau}) \| < L \| x_{\tau+1} - x_{\tau} \|$, so that we can obtain the following inequality:
\begin{equation}
f(x_0) - f(x^*) < n \cdot L \cdot \| x_{\tau+1} - x_{\tau} \| \quad \because \tau \in \mathbf{Z}^+.    
\end{equation}
Thus, we can partition the $Q_p$ as follows:  
\begin{equation}
  Q_p^{-1}
  = (f(x_0) - f(x^*)) \cdot b^{\bar{h}(t) + \epsilon} < \| x_{\tau + 1} - x_{\tau} \| \cdot n \cdot L \cdot b^{\bar{h}(t) + \epsilon}
  = \| x_{\tau + 1} - x_{\tau} \| \cdot \bar{Q}_p^{-1}
\end{equation}
, where $\bar{Q}_p^{-1} \triangleq C_0 \cdot b^{\bar{h}(t) + \epsilon}$ and $C_0 = n L$.
From the fact that $\| x_{\tau + 1} - x_{\tau} \| > \| x_{t + k + 1} - x_{t+k} \|, \; \forall k \in \mathbf{Z}[1, n]$, we can establish a positive value $p > 1$ such that
\begin{equation}
  \| x_{\tau + 1} - x_{\tau} \| = p \cdot \| x_{t + 1} - x_{t} \|
\end{equation}
, for an arbitrary $t > 0$.
By $\eqref{a2_eq01}$ and $\eqref{a2_eq02}$, we get
\begin{equation}
  f(x_{t+1}) - f(x_t) = - (\varepsilon_{t+1} - \varepsilon_t)Q_p^{-1}    
\end{equation}
and
\begin{equation}
f(x_{t+1}) - f(x_t) = -\frac{f(x_{t+1}) - f(x_t)}{\| x_{t+1} - x_t \|} v_t \cdot (x_{t+1} - x_t).    
\end{equation}
Therefore,
\begin{equation}
\begin{aligned}
f(x_{t+1}) - f(x_t)
&= - (\varepsilon_{t+1} - \varepsilon_t)Q_p^{-1} \\
&= - (\varepsilon_{t+1} - \varepsilon_t) \cdot \| x_{\tau + 1} - x_{\tau} \| \cdot \bar{Q}_p^{-1} \\
&= - (\varepsilon_{t+1} - \varepsilon_t) \cdot \| x_{\tau + 1} - x_{\tau} \| \cdot v_t \cdot v_t \cdot \bar{Q}_p^{-1} \\
&= (\varepsilon_{t+1} - \varepsilon_t) \cdot \| x_{\tau + 1} - x_{\tau} \| \cdot \frac{(x_{t + 1} - x_{t})}{\| x_{t + 1} - x_{t} \|} \cdot v_t \cdot \bar{Q}_p^{-1} \\
&= (\varepsilon_{t+1} - \varepsilon_t) \cdot p \cdot \| x_{t + 1} - x_{t} \| \cdot \frac{(x_{t + 1} - x_{t})}{\| x_{t + 1} - x_{t} \|} \cdot v_t \cdot \bar{Q}_p^{-1} \\
&= (\varepsilon_{t+1} - \varepsilon_t) \cdot (x_{t + 1} - x_{t}) \cdot v_t \cdot \tilde{Q}_p^{-1}
\end{aligned}
\end{equation}
, where $\tilde{Q}_p^{-1} = p\, C_0\, b^{\bar{h}(t) + \epsilon} = p \, \bar{Q}_p^{-1}$.
Consequently, we can obtain
\begin{equation}
\bar{\varepsilon}_{t} Q_p^{-1} = - (\varepsilon_{t+1} - \varepsilon_t) \cdot (x_{t + 1} - x_{t}) \cdot v_t \cdot \tilde{Q}_p^{-1}
\label{a3_eq02}
\end{equation}
, where $\bar{\varepsilon_t} \triangleq \varepsilon_{t+1} - \varepsilon_{t}$.
\end{proof}

\subsection{Proof of theorem 5}
\begin{theorem-ack}
For a given objective function $f(x_t) \in \mathbf{R}$, suppose that there exist the quantized objective functions $f^Q(x_t), \; f^Q(x_{t+1})$ at a current state $x_t$ and the following state $x_{t+1}$ such that $f^Q(x_t) \geq f^Q(x_{t+1})$, for all $x_{t+1} \neq x_t$; we can obtain the differential equation of the state transition as follows:
\begin{equation}
dX_t = - \nabla_x f(X_t) dt + \sqrt{C_q} \cdot Q_p^{-1}(t) dW_t
\end{equation}
where $W_t \in \mathbf{R}^n$ represents a standard Wiener process, which has a zero mean and variance with one, $X_t \in \mathbf{R}^n$ denotes a random variable corresponding to $x_t \in \mathbf{R}^n$, and $C_q \in \mathbf{R}$ is a constant value.
\end{theorem-ack}

\begin{proof}           
By Definition $\ref{def_01}$, we can write the quantized objective function $f^Q(x_t)$ as follows:
\begin{equation}
    f^Q(x_t) = f(x_t) + \varepsilon_t \cdot Q_p^{-1} (t).
\label{prop_01pf_eq01}
\end{equation}
According to $\eqref{prop_01pf_eq01}$, the difference of the quantized objective function represents
\begin{equation}
f^Q(x_{t+1}) - f^Q(x_t) = f(x_{t+1}) - f(x_t) + (\varepsilon_{t+1} - \varepsilon_t) \cdot Q_p^{-1} (t).
\label{prop_01pf_eq02}
\end{equation}
By the definition of Taylor expansion, we get
\begin{equation}
    f(x_{t+1}) - f(x_t) = \nabla_x f(x_t)(x_{t+1} - x_t) + \int_0^1 (1 - s) \frac{\partial^2 f}{\partial x^2}(x_t + s(x_{t+1} - x_t))(x_{t+1} - x_t)^2 ds.
\label{prop_01pf_eq03}
\end{equation}
From $\eqref{prop_01pf_eq02}$, we can evaluate the difference of the objective function when $f^Q(x_{t+1}) = f^Q(x_t)$ such that 
\begin{equation}
0 = f^Q(x_{t+1}) - f^Q(x_t) = f(x_{t+1}) - f(x_t) + \bar{\varepsilon}_t \cdot Q_p^{-1} (t)
\Rightarrow f(x_{t+1}) - f(x_t) = -\bar{\varepsilon}_t \cdot Q_p^{-1} (t)
\end{equation}
, where $\bar{\varepsilon}_t$ denotes a difference of $\varepsilon_{t}$ such that $\bar{\varepsilon}_t \triangleq \varepsilon_{t+1} - \varepsilon_t$ and $\bar{\varepsilon}_t \in \mathbf{R}[-1, 1]$.
Following Lipschitz continuous condition (Assumption $\ref{assum02}$) and $f^Q(x_{t+1}) - f^Q(x_t) \leq 0$, there exists a positive value $m \in \mathbf{R}^{+}$ such that
\begin{equation}
    m \triangleq \inf_{x} \left| \frac{\partial^2 f}{\partial x^2} (x)  \right|, \forall x=x_t + s(x_{t+1} - x_t), \; s \in \mathbf{R}[0,1].
\label{prop_01pf_eq04}
\end{equation}
Using $\eqref{a3_eq02}$ and $\eqref{prop_01pf_eq04}$, we rewrite $\eqref{prop_01pf_eq03}$ as follows:
\begin{equation}
\begin{aligned}
0 \geq f(x_{t+1}) - f(x_t)
&> \nabla_x f(x_t)(x_{t+1} - x_t) + m (x_{t+1} - x_t)^2 \int_0^1 (1 - s) ds + \bar{\varepsilon}_t Q_p^{-1} (t) \\
&= (x_{t+1} - x_t) \cdot \nabla_x f(x_t) + \frac{m}{2} (x_{t+1} - x_t)^2 + \bar{\varepsilon}_t Q_p^{-1} (t) \\
&= - (x_{t+1} - x_t) \cdot \left( -\nabla_x f(x_t) + v_t \cdot \bar{\varepsilon}_t \tilde{Q}_p^{-1} (t) \right) + \frac{m}{2} (x_{t+1} - x_t)^2
\end{aligned}
\label{prop_01pf_eq05}
\end{equation}
, where $v_t$ denotes a normalized vector such that $v_t = -\frac{(x_{t+1} - x_t)}{\| x_{t+1} - x_t \|}$.

In $\eqref{prop_01pf_eq05}$, if we choose $(x_{t+1} - x_t)$ appropriately, we note that there exist a positive $m$ satisfying the inequality condition $f(x_{t+1}) \leq f(x_t)$.
Thereby, when we set $x_{t+1} - x_t$ as follows
\begin{equation}
    x_{t+1} - x_t = -\nabla_x f(x_t) + v_t \cdot \bar{\varepsilon}_t \tilde{Q}_p^{-1} (t)
\label{prop_01pf_eq06}
\end{equation}
, we can obtain the following inequality:
\begin{equation}
    0 \geq f(x_{t+1}) - f(x_t) > (x_{t+1} - x_t)^2 \left( \frac{m}{2} - 1\right).
\label{prop_01pf_eq07}
\end{equation}
Consequently, when the infimum to the second derivation of the objective function $f(x)$ fulfills $0 \leq m < 2$, we can find the state $x_{t+1}$ satisfying the inequality $f(x_{t+1}) - f(x_t)$.
Conversely, if $m \geq 2$, it contradicts $f(x_{t+1}) \leq f(x_t)$. In other words, $\eqref{prop_01pf_eq07}$ turns to the follwoing inequality:
\begin{equation}
f(x_{t+1}) - f(x_t) > (x_{t+1} - x_t)^2 \left( \frac{m}{2} - 1\right) \geq 0, \; \forall m \geq 2.
\label{prop_01pf_eq07-01}
\end{equation}
$\eqref{prop_01pf_eq07-01}$ implies $f(x_{t+1}) > f(x_t)$, and it means that the proposed algorithm brings a hill-climbing effect within the domain fillfills the quantized range such as $x \in \{ x | f^Q(x) = f^Q(x_{t+1}) = f^Q(x_t) \}$.
Since the proposed algorithm serves the bounded range provided by the quantization for each iteration, the hill-climbing effect cannot lead to divergence of the algorithm.

To obtain a differential form of the difference to $x_t$, we let $Z(s) = x_t + s (x_{t+1} - x_t )$ and rewrite $\eqref{prop_01pf_eq06}$ as following integral equation:
\begin{equation}
    x_{t+1} - x_t = (x_{t+1} - x_t)\int_0^1 ds = \int_0^1 (x_{t+1} - x_t) ds = \int_0^1 \frac{\partial Z(s)}{\partial s} ds = \int_0^1 dZ(s).
\label{prop_01pf_eq08}    
\end{equation}
From $\eqref{prop_01pf_eq06}$, we can get
\begin{equation}
    x_{t+1} - x_t = \int_0^1 dZ(s) = \int_0^1 (x_{t+1} - x_t) ds = \int_0^1 (-\nabla_x f(x_t) + v_t \cdot \bar{\varepsilon}_t \tilde{Q}_p^{-1} (t)) ds = \int_t^{t+1} dx_{s}.
\label{prop_01pf_eq09}    
\end{equation}

Herein, since $v_t$ is a normalized vector, we can get the variance of $v_t \cdot \bar{\varepsilon}_t \tilde{Q}_p^{-1}(t)$ such that
\begin{equation}
    \mathbb{E}_{\mathcal{F}_t} \langle v_t, v_t \rangle \cdot \bar{\varepsilon}^2_t \tilde{Q}_p^{-2}(t)
    =  \tilde{Q}_p^{-2}(t) \mathbb{E}_{\mathcal{F}_t} \bar{\varepsilon}^2_t
    = \frac{4}{12 \cdot \tilde{Q}_p^2 (t) }=C_q \tilde{Q}_p^{-2} (t)\quad \because \|v_t \| = 1,\; C_q = 1/3
\end{equation}
Differentiating the two right-most terms in $\eqref{prop_01pf_eq09}$, we obtain
\begin{equation}
\begin{aligned}
    \frac{\partial}{\partial s} \int dx_{s} \bigg\vert_{s=t}
    &= \frac{\partial }{\partial s} \int (-\nabla_x f(x_t) + v_t \cdot \bar{\varepsilon}_t \tilde{Q}_p^{-1} (t)) ds \bigg\vert_{s=t} \\
\implies    dX_t &= -\nabla_x f(X_t) dt + v_t \cdot \bar{\varepsilon}_t \tilde{Q}_p^{-1} (t) dt \\
\implies    dX_t &= -\nabla_x f(X_t) dt + \sqrt{C_q} \cdot \tilde{Q}_p^{-1} dW_t
\end{aligned}    
\end{equation}
\end{proof}
Theorem $\ref{th_005}$ gives the fundamental stochastic differential form to evaluate the optimal quantization schedule for global optimization.

In $\eqref{prop_01pf_eq07}$, you can argue that, if $m$ is larger than two, then the inequality is broken.
However, since $m$ is just an infimum of the second derivation of the objective function, not the correct value, we can regard it as a quadratic approximated function to the objective function.
Thus, the proposition holds sufficiently when the objective function is locally convex on some domain around $x_t$.
The more important point is that the proposition also holds when  $m$ is negative or zero. 
Negative $m$ is that the objective function is a concave function on a local domain of $x_t$. 
In a conventional convex optimization theory, we cannot obtain a less value of an objective function at a next state $x_{t+1}$ based on a negative gradient than a current value.
However, since the proposed algorithm can get a lower objective function value at the next state despite a concave function, $m$ can be equal to or less than zero.    
Additionally, when the value of the objective function on the next state is larger than the current state, the quantization makes the next and the current value of the objective function equal so that the proposition still holds. 

Even though the proposed algorithm does not have any scheduler similar to the temperature scheduler in simulated annealing, if the quantization parameter decreases to the schedule provided by the following proposition, the proposed algorithm can find the global optimum.
\section{Proof of Theorems in Section 4}
\subsection{Proof of theorem 6}
\begin{theorem-ack}
If the dynamics of the state transition by the proposed algorithm follow $\eqref{eq_th05_01}$, the state $x_t$ weakly converges to the global minimum when the quantization parameter decreases to the following schedule:
\begin{equation}
\inf_{t \geq 0} Q_p^{-1}(t) = \frac{C_o}{\log (t + 2)}, \quad C_o \in \mathbf{R}^+,\; C_o \gg 0
\end{equation}
\end{theorem-ack}

\begin{proof}       
For the proof of the theorem, we depend on the lemmas in works of \citet{Geman-1986}.
First, we prove the following convergence of the transition probability:
\begin{equation}
\lim_{\tau \rightarrow \infty} \sup_{x_t, x_{t+\tau} \in \mathbf{R}^n} \| p(t, \bar{x}_t, t + \tau,  x^*) - p(t, x_t, t + \tau,  x^*) \| = 0
\label{lmpf2.7_eq01}    
\end{equation}
, where $t$ and $\tau$ is the current time index and the process time index, respectively. $x^*$ represents an global optimum for the objective function $f(x_t)$.

Let the infimum of the transition probability from $t$ to $t+1$ such that
\begin{equation}
\delta_t = \inf_{x, y \in \mathbf{R}^n} p(t, x, t+1, y) 
\label{lmpf2.7_eq02}    
\end{equation}
According to the lemma in \citet{Geman-1986}, we can evaluate the upper bound of $\eqref{lmpf2.7_eq01}$ as follows:

\begin{equation}
\begin{aligned}
&\overline{\lim_{t \rightarrow \infty}} \sup_{v, w} \vert p(s,v, t, f) - p(s, w, t, f) \vert \\
&= \overline{\lim_{t \rightarrow \infty}} \sup_{v, w} 
\left\vert \int p(s,v, s+1, z)p(s+1,z, t, f) dz
- \int p(s, w, s+1, z) p(s+1, z, t, f) dz \right\vert \\
&= \overline{\lim_{t \rightarrow \infty}} \sup_{v, w} 
\left\vert \int p(s,v, s+1, z)  p(s+1,z, t, f) dz
- \int p(s, w, s+1, z) p(s+1, z, t, f) dz - (\delta_s - \delta_s) p(s+1,z, t, f) \right\vert \\
&= \overline{\lim_{t \rightarrow \infty}} \sup_{v, w} 
\left\vert \int (p(s,v, s+1, z) - \delta_s) p(s+1,z, t, f) dz - \int (p(s, w, s+1, z) - \delta_s) p(s+1, z, t, f) dz \right\vert  \\
&\leq \overline{\lim_{t \rightarrow \infty}} \sup_{v, w} 
\left\vert \int (p(s,v, s+1, z) - \delta_s) \sup_{z} p(s+1,z, t, f) dz 
- \int (p(s, w, s+1, z) - \delta_s) \inf_{z} p(s+1, z, t, f) dz \right\vert \\
&= \overline{\lim_{t \rightarrow \infty}} \sup_{v, w} 
\left\vert \sup_{z} p(s+1,z, t, f) \int (p(s,v, s+1, z) - \delta_s) dz
- \inf_{z} p(s+1, z, t, f) \int ( p(s, w, s+1, z) - \delta_s) dz \right\vert \\
&\leq \overline{\lim_{t \rightarrow \infty}} \sup_{v, w} 
\left\vert ( 1 - \delta_s) \sup_{z} p(s+1,z, t, f) - (1 - \delta_s) \inf_{z} p(s+1,z, t, f) \right\vert \\
&= \overline{\lim_{t \rightarrow \infty}} 
\sup_{v, w} ( 1 - \delta_s) \left\vert \sup_{z} p(s+1,z, t, f) - \inf_{z} p(s+1,z, t, f) \right\vert \\
&\cdots \\
&\leq \overline{\lim_{t \rightarrow \infty}}
\left(\prod_{k=0}^{(t-s)-1} (1 - \delta_{s+k}) \right) \cdot
\sup_{v, w} \left\vert p(s+(t-s),v, t, f) - p(s+(t-s), w, t, f) \right\vert \\
&\leq \overline{\lim_{t \rightarrow \infty}} \left(\prod_{k=0}^{(t-s)-1} (1 - \delta_{s+k}) \right)
= \prod_{k=0}^{\infty}(1 - \delta_{s+k}).
\end{aligned}
\end{equation}
Thus, we obtain
\begin{equation}
\overline{\lim_{\tau \rightarrow \infty}} \sup_{x_t, x_{t+\tau} \in \mathbf{R}^n} \| p(t, \bar{x}_t, t + \tau,  x^*) - p(t, x_t, t + \tau,  x^*) \| \leq \prod_{k=0}^\infty (1 - \delta_{t+k}).
\label{lmpf2.7_eq03}    
\end{equation}
From the exponential approximation \eqref{eq01:lemma} in Lemma:Auxiliary,  we rewrite $\eqref{lmpf2.7_eq03}$ as follows:
\begin{equation}
\overline{\lim_{\tau \rightarrow \infty}} \sup_{x_t, x_{t+\tau} \in \mathbf{R}^n} \| p(t, \bar{x}_t, t + \tau,  x^*) - p(t, x_t, t + \tau,  x^*) \| \leq \exp(-\sum_{k=0}^{\infty} \delta_{t+k}) ).
\label{lmpf2.7_eq04}    
\end{equation}
Herein, to obtain the bound of $\delta_{t+k}$, we rewrite the SDE for the dynamics of the proposed algorithm from Theorem $\ref{th_006}$:
\begin{equation}
dX_s = - \nabla f(X_{s}) ds + \sigma(s) \sqrt{C_q} dW_s, \quad s \in \mathbf{R}(t, t+1).
\label{lmpf2.7_eq05}    
\end{equation}
, where $\sigma(s) \triangleq Q_p^{-1}(s)$.

Define a domain $\mathcal{F} \{ f: [t, t+1] \rightarrow \mathbf{R}^n, f \text{ continuous } \}$, 
Let $P_x$ be the probability measures on $\mathcal{F}$ induced by $\eqref {lmpf2.7_eq05}$ and the probability distribution $Q_x$ given by the following equation:
\begin{equation}
d\bar{X}_{s} = \sigma(s) \sqrt{C_q} dW_{s}, \quad s \in \mathbf{R}(t, t+1). 
\label{lmpf2.7_eq06}    
\end{equation}
According to the Girsanov theorem (\citet{Bernt_2003, Klebaner_2011}), we obtain
\begin{equation}
\frac{dP_{x}}{dQ_{x}}
= \exp \left\{-\int_t^{t+1} \frac{C_q^{-1}}{\sigma^2 (s)} \nabla_x f(X_{s}) d\bar{X}_{s} -\frac{1}{2} \int_t^{t+1} \frac{C_q^{-1}}{\sigma^2 (s)} \|  \nabla_x f(X_{s}) \|^2 ds \right\}.
\label{lmpf2.7_eq07}    
\end{equation}

To compute the upper bound of $\eqref{lmpf2.7_eq07}$, we will check the upper bound of $\| \nabla_x f \|$.  
Considering Assumption $\ref{assum02}$, the gradient of $f(x_t) \in C^{2}$ fulfills the Lipschitz continuous condition as well. 
Thereby, there exist a positive value $L'$  such that 
\begin{equation}
\| \nabla f(w_{s}) - \nabla f(x^*) \| \leq L' \| w_{s} - x^* \|, \quad \forall s > 0.
\label{lmpf2.7_eq09}    
\end{equation}
Successively, since $\nabla_x f(x^*) = 0$, the Lipschitz condition forms simply as follows : 
\begin{equation}
\| \nabla_x f(x_t) \| \leq L' \rho = C_0
\label{lmpf2.7_eq10}    
\end{equation}
, where $\rho = \| x_t - x^* \|$.

Consequently, for all $s \in \mathbf{R}[t, t+1)$, we compute the upper bound of the first term in exponential function in $\eqref{lmpf2.7_eq07}$ as follows:
\begin{equation}
\begin{aligned}
\left\| \int_t^{t+1} \frac{C_q^{-1}}{\sigma^2 (s) } \nabla_x f(X_{s}) d\bar{X}_{s} \right\| 
&\leq  \int_t^{t+1} \left\| \frac{C_q^{-1}}{\sigma^2 (s)} \nabla_x f(X_{s}) d\bar{X}_{s} \right\| \\
&\leq  \int_t^{t+1} \frac{ C_q^{-1}}{\sigma^2 (s)} \left\| \nabla_x f(X_{s}) \right\| \sigma (s) \sqrt{C_q} d W_{s}    \\
&\leq  \frac{\sqrt{ C_q^{-1}}}{\sigma (s)} \sup \left\| \nabla_x f(X_{s}) \right\|  \int_{t}^{t+1}  dW_s \\
&\leq  \frac{\sqrt{C_q^{-1}}}{\sigma (s)} C_0  \| W_t - \frac{1}{2} \|  
\leq \frac{1}{\sigma (s)} C_0 \sqrt{ C_q^{-1}} (\rho + \frac{1}{2}). 
\end{aligned}
\label{lmpf2.7_eq14}    
\end{equation}
$\eqref{lmpf2.7_eq14}$ implies that 
\begin{equation}
\left\| -\int_t^{t+1} \frac{C_q^{-1}}{\sigma (s) } \nabla_x f(X_{s}) d \bar{X}_{s} \right\| \leq \frac{C_1}{\sigma (s)}
\label{lmpf2.7_eq15}    
\end{equation}
, where $C_1$ denotes positive value such that $C_1 > C_0 \sqrt{C_q^{-1}} (\rho + \frac{1}{2}) $.

In addition, the upper bound of the second term represents  
\begin{equation}
\begin{aligned}
\frac{1}{2} \left\| \int_t^{t+1} \frac{C_q^{-1}}{\sigma^2 (s) } \|  \nabla_x f(X_{s}) \|^2 ds \right\|  
&\leq \frac{1}{2} \int_t^{t+1} \frac{ C_q^{-1} }{\sigma^2 (s)} \|  \nabla_x f(X_{s}) \|^2 ds \\
&\leq \frac{1}{2} \frac{ C_q^{-1} }{\sigma^2 (s)} \sup \|  \nabla_x f(X_{s}) \|^2 \int_t^{t+1} ds \\
&\leq \frac{1}{2 \sigma^2 (s)} C_q^{-1} \cdot C_0^2 
\leq \frac{C_2}{2 \sigma^2 (s)},   \quad \because C_2 > C_q^{-1} \cdot C_0^2. 
\end{aligned}
\label{lmpf2.7_eq16}
\end{equation}

Because $\sigma(s) \triangleq Q_p^{-1}(t)$ is monotone decreasing function, the supremum of  $\sigma(s)$ is $\sigma(0) $ for all $s \in \mathbf{R}[0, \infty)$, i.e. $\sup_{s \in \mathbf{R}[0, \infty]}  \sigma(s) = \sigma (0) \triangleq \sigma$.
With the supremum of each term in $\eqref{lmpf2.7_eq07}$, we can obtain the lower bound of the Radon-Nykodym derivative $\eqref{lmpf2.7_eq07}$  such that
\begin{equation}
\frac{dP_{w}}{dQ_{w}} \geq \exp \left( - \frac{1}{\sigma (s)}\left( C_1 + \frac{C_2}{2 \sigma (s)} \right)\right) \geq \exp \left(- \frac{C_3}{\sigma (s)}  \right), \quad \because C_3 > 2 \sigma(0) C_2 + C_1.
\label{lmpf2.7_eq17}    
\end{equation}
Accordingly, for any $\varepsilon > 0$  and $x_t, \; x^* \in \mathbf{R}^n$,  the infimum of $P_x (|X_{t+1} - x^*| < \varepsilon) $ is 
\begin{equation}
P_x (|X_{t+1} - x^*| < \varepsilon) 
\geq \exp\left(- \frac{C_3}{\sigma(s)} \right) Q_x (|X_{t+1} - x^*| < \varepsilon).
\label{lmpf2.7_eq18}    
\end{equation}
As $Q_w$ is a normal distribution based on $\eqref{lmpf2.7_eq06}$,  we have
\begin{equation}
\begin{aligned}
P_x (|X_{t+1} - x^*| < \varepsilon) 
&\geq \exp\left(- \frac{C_3}{\sigma (s)} \right) \int_{\| x - x^* \| < \varepsilon} \frac{1}{\sigma \sqrt{2 \pi \int_{t}^{t+1}C_q d\tau}} \exp \left( -\frac{ (x - x^*)^2}{2 \int_{t}^{t+1} C_q d\tau}  \right) dx \\
&\geq \exp\left(- \frac{C_3}{\sigma (s)} \right) \int_{\| x - x^* \| < \varepsilon} \frac{1}{\sigma \sqrt{2 \pi C_q \int_{t}^{t+1} d\tau}} \exp \left( -\frac{ (\sqrt{\rho} + \varepsilon)^2}{2  C_q \int_{t}^{t+1} d\tau}  \right) dx  \\
&\geq \exp\left(- \frac{C_3}{\sigma (s)} \right) \frac{1}{\sigma(0) \sqrt{2 \pi C_q}} \exp \left( -\frac{ (\sqrt{\rho} + \varepsilon)^2}{2  C_q } \right) \int_{\| x - x^* \| < \varepsilon} dx \\
&= \exp\left(- \frac{C_3}{\sigma (s)} \right) \frac{1}{\sigma(0) \sqrt{2 \pi C_q}} \exp \left( -\frac{ (\sqrt{\rho} + \varepsilon)^2}{2  C_q } \right) 2\varepsilon \\
&\geq \exp\left(- \frac{C_3}{\sigma (s)} \right) \frac{1}{\sigma(0) \sqrt{2 \pi C_q}} \left(1 + \frac{ (\sqrt{\rho} + \varepsilon)^2}{2  C_q } \right) 2\varepsilon \\
&\geq \exp\left(- \frac{C_3}{\sigma (s)} \right) \frac{1}{\sigma(0) \sqrt{2 \pi C_q}} \left(\frac{2  C_q+ (\sqrt{\rho} + \varepsilon)^2}{  C_q } \right) \Bigg\vert_{\rho=0, \varepsilon=0} \cdot \varepsilon \\
&\geq \exp\left(- \frac{C_3}{\sigma (s)} \right) \cdot C_4 \cdot \varepsilon, \quad \because C_4 = \frac{\sqrt{2}}{\sigma(0) \sqrt{\pi C_q}}. 
\end{aligned}
\label{lmpf2.7_eq19}    
\end{equation}

Finally, we obtain the lower bound of the transition probability  such that 
\begin{equation}
\label{lmpf2.7_eq20}
\begin{aligned}
\delta_t 
&= \inf_{x, y \in \mathbf{R}^n} p(t, x, t+1, y) \bigg\vert_{x = x_t,\; y=x^*} \nonumber\\
&= \inf_{x, y \in \mathbf{R}^n} \lim_{\varepsilon \rightarrow 0} \frac{1}{\varepsilon} P_x (|X_{t+1} - x^*| < \varepsilon) \\
&\geq \inf_{x, y \in \mathbf{R}^n} \lim_{\varepsilon \rightarrow 0} \frac{1}{\varepsilon} \cdot C_4 \cdot \exp\left(- \frac{C_3}{\sigma (t)} \right) \cdot  \cdot \varepsilon \\
&\geq \exp \left(-\frac{C_5}{\sigma(t)}  \right), \quad \because C_5 > C_3 + \sigma(0) \cdot | \ln C_4 |
\end{aligned}    
\end{equation}
The above inequality implies that, if there exists a monotone decreasing function such that $\sigma(s) \geq \frac{C_5}{ \log (t+2)}$, it satisfies that the convergence condition given by $\eqref{lmpf2.7_eq04}$ such that
\begin{equation}
\sum_{k=0}^{\infty} \delta_{t+k} 
\geq \sum_{k=0}^{\infty} \exp \left( -\frac{C_5}{C_5} \log(t+2+k) \right) 
= \sum_{k=0}^{\infty} \frac{1}{t+2+k} 
= \infty, \quad \forall k \geq 0.    
\label{lmpf2.7_eq21}
\end{equation}
Substitute $\eqref{lmpf2.7_eq21}$ into $\eqref{lmpf2.7_eq04}$, we obtain
\begin{equation}
    \overline{\lim_{\tau \rightarrow \infty}} \sup_{x_t, x_{t+\tau} \in \mathbf{R}^n} \| p(t, \bar{x}_t, t + \tau,  x^*) - p(t, x_t, t + \tau,  x^*) \| \leq 2 \| x^* \|_{\infty} \exp(-\sum_{k=0}^{\infty} \delta_{t+k}) ) = 0.
\end{equation}
\end{proof}

\subsection{Proof of theorem 7}
Although Theorem $\ref{th_006}$ provides the scheduler of the quantization parameter obtaining the global minimum, the scheduler is not practical. 
Whereas the quantization parameter is a rational number, the value of the scheduler is a real number.
Therefore, we have to set an appropriate bound of the scheduler for the quantization parameter. 
The following theorem gives one instance. 

\begin{theorem-ack}
Suppose that there exists an integer valued annealing schedule $\sigma(t) \in \mathbf{Z}^+$ such that $\sigma(t) \geq \inf \sigma(t) \triangleq c/\log(t+2)$.  If the power function $\bar{h}(t)$ of the quantization parameter $Q_p^{-1}(t)$ fulfills the following condition, the proposed algorithm weakly converges to the global optimum.
\begin{equation}
    \log_b \left(C_0 \cdot b^{-\frac{2 \beta}{t+2}} \cdot \inf \sigma(t) \right) \leq \bar{h}(t) \leq \log_b \left( C_1 \log(t+2) \right)
\end{equation}
, where $C_0 \equiv \eta \sqrt{C_q}$ and $C_1 \equiv \sqrt{C_q}\eta/C$.
\end{theorem-ack}
\begin{proof}       
From the Theorem $\ref{th_005}, \ref{th_006}$, we obtain the infimum of $\sigma(t) \triangleq \sqrt{C_q} Q_p^{-1}(t)$.
To evaluate the integer value of the quantization resolution $Q_p(t)$, we set $T(t)$ to be a supremum of $\sigma(t)$ such that
\begin{equation}
    \frac{C}{\log (t+2)} \leq \sigma(t) \leq T(t).
\label{pf4_3_eq01}
\end{equation}
In $\eqref{pf4_3_eq01}$, $T(t)$ is a monotone decreasing function, such as $T(t) \downarrow 0$ with respect to $ t \uparrow 0$.
Moreover, when $\Delta$ is given as $\Delta \equiv \sup_{x,y \in \mathbf{R}} (f(x) - f(y)$, $T(t)$ includes the following properties:
\begin{equation}
    \frac{d}{dt} \exp \left( -\frac{2 \Delta}{T(t)}\right) 
    = \frac{dT(t)}{dt} \cdot \frac{1}{T^2(t)} \exp \left( -\frac{2 \Delta}{T(t)}\right) \rightarrow 0, 
    \quad \text{as}\; t \uparrow \infty
\label{pf4_3_eq02}    
\end{equation}
From Definition $\ref{def_02}$, we note $Q_p = \eta \cdot b^{-\bar{h}(t)}$, so that we substitute $\sigma(t)$ with $Q_p(t)$ in $\eqref{pf4_3_eq01}$, as follows:
\begin{equation}
    \frac{C}{\log (t+2)} \leq \sqrt{C_q} \cdot \eta \cdot b^{-\bar{h}(t)} \leq T(t).
\label{pf4_3_eq03}    
\end{equation}
Applying the log function to each term and rearranging, we obtain 
\begin{equation}
\log_b \left( \frac{\sqrt{C_q}\eta }{T(t)}  \right) \leq {\bar{h}(t)} \leq \log_b \left( \frac{\sqrt{C_q}\eta \cdot \log (t+2)}{C} \right).    
\label{pf4_3_eq04}    
\end{equation}
Let $T(t) \triangleq b^{\frac{2 \beta}{t+2}} \cdot (\inf_{t \geq 0} \sigma(t))^{-1}$, then we get
\begin{equation}
    \log_b \left( \eta \sqrt{C_q} \cdot b^{-\frac{2 \beta}{t+2}} \inf_{t \geq 0} \sigma(t)\right) \leq {\bar{h}(t)} \leq \log_b \left( \frac{\sqrt{C_q}\eta \cdot \log (t+2)}{C} \right).
\end{equation}
Since $C_0 \equiv \eta \sqrt{C_q}$ and $C_1 \equiv \sqrt{C_q}\eta/C$, the theorem holds.  
\end{proof}
\section{Detailed Information of Simulation Results}
\begin{figure}[htb!] 
    \centering
    \subfigure[Initial path given by the nearest neighborhood algorithm (cost is 2159)]{\resizebox{0.48\textwidth}{!}{\input{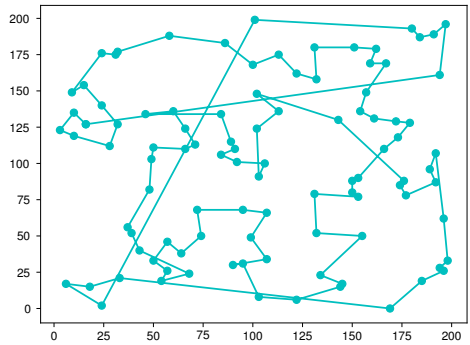}}}\label{fig01-000}
    \hfill
    \subfigure[Final path given by the simulated annealing algorithm (the minimum cost is 1731)]{\resizebox{0.48\textwidth}{!}{\input{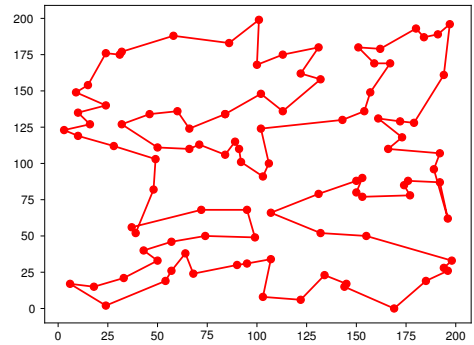}}}\label{fig01-001}
    \hfill
    \subfigure[Final path given by the quantum annealing algorithm (the minimum cost is 1706)]{\resizebox{0.48\textwidth}{!}{\input{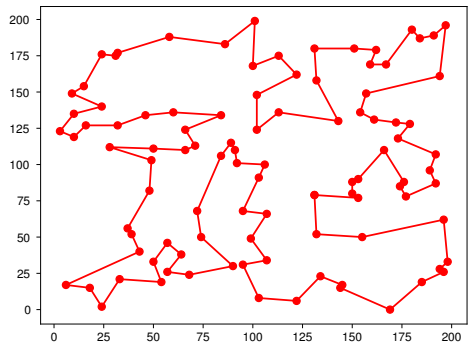}}}\label{fig01-002}
    \hfill
    \subfigure[Final path given by the proposed algorithm (the minimum cost is 
    1636)]{\resizebox{0.48\textwidth}{!}{\input{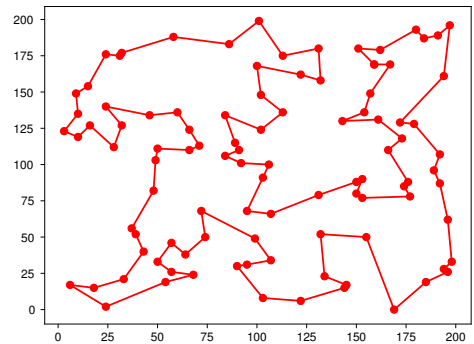}}}\label{fig01-003}
\caption{Comparison of TSP routes provided by each optimization algorithm}
\label{fig_tsp_01}
\end{figure}
\begin{figure}[htb!]      
    \centering
    \resizebox{\textwidth}{!}{
    \input{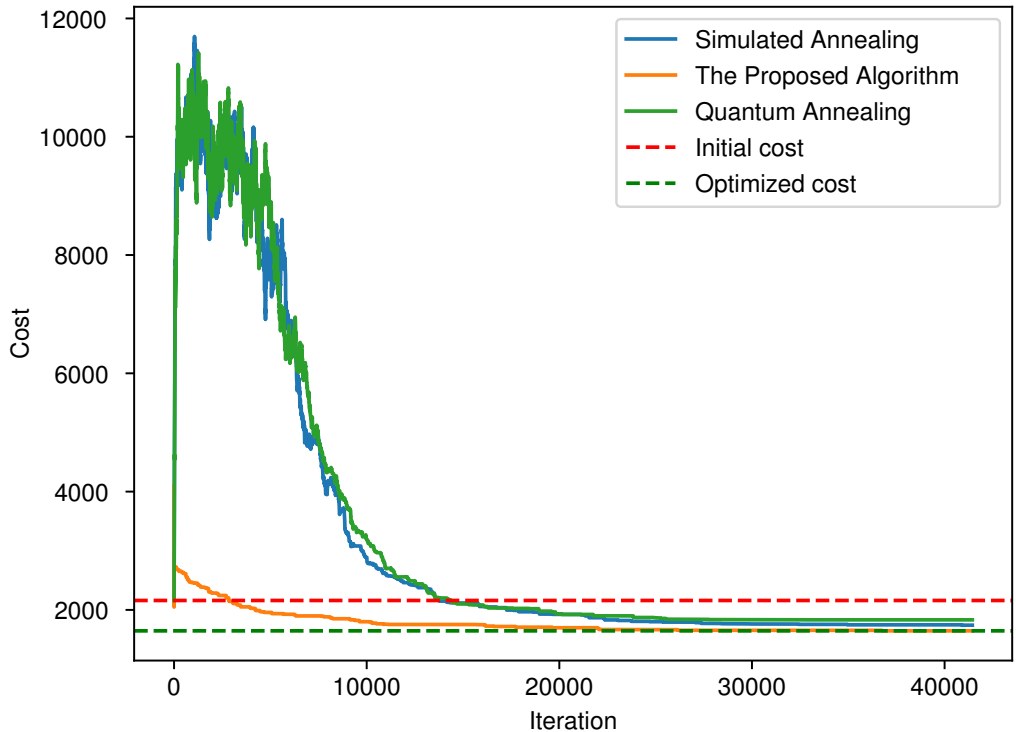}
    }
    \caption{The minimum cost trends corresponding to each algorithm to iterations in the TSP optimization test}
    \label{fig_tsp_02}
\end{figure}
As mentioned in the manuscript, we perform the optimization test with equal and fixed locations of cities for all attempts. 
Figure $\ref{fig_tsp_01}$ shows the initial path given by the nearest neighbor algorithm, the final route given by SA, QA, and the proposed optimization algorithm.

Figure $\ref{fig_tsp_02}$ shows the trends of the minimum cost produced by each tested algorithm.
Because SA and QA exploit an acceptance probability, the trends of the errors in both algorithms represent fluctuation in the early stage of optimization.
However, the proposed algorithm does not include acceptance probability, so that the minimum cost decreases with relatively small fluctuation seen in SA and QA.

In addition, the slight fluctuation given by the proposed algorithm provides a fast convergence to a feasible (or global) solution compared to other algorithms.
The quantization used in the proposed algorithm gives a hill-climbing effect as other heuristic algorithms do.
However, the proposed algorithm suppresses the hill climbing effect reasonably so that the candidates selected by the optimization algorithm cannot diverge to an unfeasible solution so far.
On the other hand, the other algorithms permit the candidate to diverge under the acceptance probability. Therefore, those algorithms require more computation time to converge, even if the algorithm can find the global minima.

The reasonable hill-climbing provided by the proposed algorithm represents robust optimization performance compared to other algorithms using an acceptance probability. 

\newpage
\bibliography{main}

\end{document}